%% file: main_mmt.tex
\documentclass[a4paper,fleqn]{cas-sc}

\usepackage[numbers,longnamesfirst]{natbib}

\usepackage{amssymb,amsmath,amsthm}
\usepackage{graphicx}
\usepackage{hyperref}
\usepackage{kbordermatrix}
\usepackage{pgfplots}
\pgfplotsset{compat=newest}

\def\tsc#1{\csdef{#1}{\textsc{\lowercase{#1}}\xspace}}
\tsc{WGM}
\tsc{QE}
\tsc{EP}
\tsc{PMS}
\tsc{BEC}
\tsc{DE}
\newcommand{\cln}{\,:\,}
\newcommand{\ve}[1]{\mathbf v(#1)}
\DeclareMathOperator{\tr}{tr}

\theoremstyle{remark}%
\newtheorem{remark}{Remark}%

\begin{document}
\let\WriteBookmarks\relax
\def\floatpagepagefraction{1}
\def\textpagefraction{.001}
\shorttitle{Forward kinematics of a general Stewart--Gough platform by elimination templates}
\shortauthors{Evgeniy Martyushev}

\title[mode = title]{Forward kinematics of a general Stewart--Gough platform by elimination templates}



\author{\color{black}{Evgeniy} Martyushev}[
                        auid=000,bioid=1,
                        orcid=0000-0002-6892-079X]
\ead{martiushevev@susu.ru}

\credit{Conceptualization, Methodology, Software, Validation, Visualization, Writing}

\affiliation{organization={Institute of Natural Sciences and Mathematics, South Ural State University},
                addressline={76 Lenin Avenue},
                city={Chelyabinsk},
                citysep={}, 
                postcode={454080},
                country={Russia}}

%

\begin{abstract}
The paper proposes an efficient algebraic solution to the problem of forward kinematics for a general Stewart--Gough platform. The problem involves determining all possible postures of a mobile platform connected to a fixed base by six legs, given the leg lengths and the internal geometries of the platform and base. The problem is known to have 40 solutions (whether real or complex). The proposed algorithm consists of three main steps: (i) a specific sparse matrix of size $293\times 362$ (the elimination template) is constructed from the coefficients of the polynomial system describing the platform's kinematics; (ii) the PLU decomposition of this matrix is used to construct a pair of $69\times 69$ matrices; (iii) all 40 solutions (including complex ones) are obtained by computing the generalized eigenvectors of this matrix pair. The proposed algorithm is numerically robust, computationally efficient, and straightforward to implement --- requiring only standard linear algebra decompositions. MATLAB, Julia, and Python implementations of the algorithm will be made publicly available.
\end{abstract}


\maketitle

\section{Introduction}
\label{sec:intro}

\begin{figure}[ht]
\label{fig:sgp}
\centering 
\includegraphics[scale=0.4]{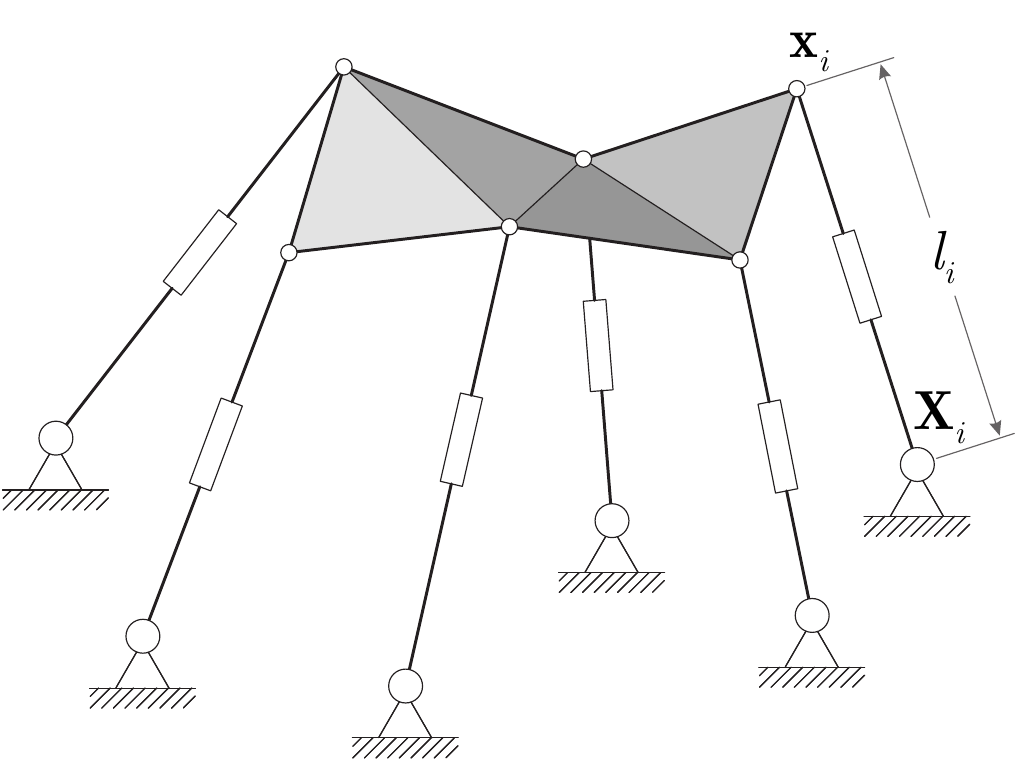}
\caption{A general Stewart--Gough platform: both the base and top platform may be non-planar and asymmetric}
\end{figure}

A Stewart--Gough platform (SGP) is a parallel manipulator widely used in applications such as flight simulators, robotic manipulators, haptic devices, surgical robots, radio telescopes, etc. It consists of a fixed base and a movable platform linked by six independently extensible legs. Each leg is attached to both the base and the platform via spherical joints, allowing for a wide range of motion, see Figure~\ref{fig:sgp}. The leg lengths are typically adjusted using actuators --- such as hydraulic or pneumatic cylinders --- allowing precise control over the platform's position and orientation~\cite{gough1962universal,stewart1965platform}.

The general SGP makes no assumptions about planarity or symmetry in either the fixed base or movable platform. This fully parallel mechanism has six degrees of freedom, meaning it can move in three translational directions and three rotational directions (pitch, roll, yaw). This allows the platform to move freely in 3-dimensional space.

The forward (or direct) kinematics problem for a Stewart--Gough platform involves determining the position and orientation of the movable platform given the lengths of the six extensible legs and the internal geometries of the platform and base. This problem is inherently complex due to the parallel structure of the platform, resulting in a system of non-linear polynomial equations. It has been shown that the problem has exactly 40 solutions in complex space~\cite{raghavan1993stewart,lazard1993representation,mourrain199340,wampler1996forward}.

Since the early 1990s, many approaches have been developed to address the forward kinematics problem for a general SGP. Some of them include the use of Newton--Raphson and Levenberg--Marquardt algorithms for local refinement~\cite{merlet2002direct,wang2007direct,grishin2023numerical}; polynomial resultant theory and Gr\"obner bases~\cite{husty1996algorithm,lee2001forward,lee2003improved,huang2009forward,nag2021uniform}; interval analysis~\cite{didrit1998guaranteed,merlet2004solving}; metaheuristic optimization strategies, such as genetic algorithms~\cite{boudreau1996solving,sheng2006forward,omran2009optimal} and particle swarm optimization~\cite{yin2022new}; precomputed lookup tables~\cite{tarokh2007real}; machine learning techniques, such as neural networks~\cite{sadjadian2005neural,parikh2009solving,chauhan2022forward} and regression~\cite{morell2013solving}. Each approach has its own strengths and limitations, see survey~\cite{safeena2022survey} for a description and comparison of some of the methods.

In addition to general approaches, specialized methods have been developed for SGPs with specific geometric configurations or augmented sensing capabilities. These approaches exploit simplified kinematics from symmetric or planar base/platform designs~\cite{griffis1989forward,innocenti1990direct,nielsen1996direct,ji2001closed,huang2007forward}, additional sensor data to reduce solution space~\cite{schulz2017sensor,seibel2018direct,karmakar2024non}. Such adaptations often yield closed-form solutions or order-of-magnitude speed improvements compared to general-case algorithms.

While substantial progress has been made in developing forward kinematics solvers for a general SGP, achieving an optimal balance between computational efficiency, numerical accuracy, and implementation simplicity remains an open fundamental challenge in parallel robotics. This paper addresses this challenge through a novel algebraic solver based on the method of elimination templates. This method, widely adopted in computer vision for minimal problems since the late 2000s~\cite{kukelova2008automatic,larsson2017efficient,larsson2018beyond,bhayani2020sparse,martyushev2022optimizing,martyushev2024automatic}, provides the following advantages for the problem of forward kinematics:
\begin{itemize}
\item computation of all 40 solutions (real and complex);
\item simultaneous maintenance of numerical accuracy and computational efficiency;
\item reliance on standard linear algebra decompositions (PLU and QZ);
\item extension to two special SGP cases without modification (Section~\ref{sec:exper}).
\end{itemize}

The rest of the paper is organized as follows. Section~\ref{sec:templ} briefly introduces the method of elimination templates for solving 0-dimensional systems of polynomial equations. Section~\ref{sec:prob} formulates the forward kinematics problem as a system of six polynomial equations. Section~\ref{sec:alg} details the proposed elimination template based algorithm for solving the forward kinematics problem for a general SGP. Section~\ref{sec:exper} validates the solver in a series of experiments on synthetic data, evaluating in particular the numerical accuracy and computational performance of the solver. Finally, Section~\ref{sec:concl} discusses the results of the paper. For complete reproducibility, the Appendix includes all necessary data required to construct the elimination template.

\section{Elimination templates}
\label{sec:templ}

In this section we briefly recall the method of elimination templates for efficiently finding roots of 0-dimensional polynomial systems, see~\cite{kukelova2008automatic,byrod2009fast,larsson2017efficient,larsson2018beyond,bhayani2020sparse,martyushev2022optimizing,martyushev2024automatic} for details.

Let $X = \{x_1, \ldots, x_k\}$ be a set of $k$ variables, $\mathbb C[X]$ the polynomial ring over the complex field $\mathbb C$, and
\begin{equation}
\label{eq:U}
U = \{x_1^{\alpha_1}\ldots x_k^{\alpha_k} \cln \alpha_i \in \mathbb Z_+\}
\end{equation}
the set of monomials in $X$. We denote by $\ve{\mathcal A}$ the vector consisting of the elements of a finite set of monomials $\mathcal A \subset U$ which are ordered according to a certain total ordering on $U$.

Given a polynomial $f \in \mathbb C[X]$, we denote by $U_f$ the support of $f$, i.e.,
\begin{equation}
\label{eq:Uf}
U_f = \{m \in U \cln c(f, m) \neq 0\},
\end{equation}
where $c(f, m)$ is the coefficient of $f$ at monomial $m$. Given a set of polynomials $F = \{f_1, \ldots, f_s\} \subset \mathbb C[X]$, we denote by $U_F$ the support of $F$, that is
\begin{equation}
\label{eq:UF}
U_F = \bigcup_{i = 1}^s U_{f_i}.
\end{equation}
Let $n = \# U_F$ be the cardinality of the finite set $U_F$. The \emph{Macaulay matrix} $M(F) \in \mathbb C^{s\times n}$ is defined as follows: its $(i, j)$th element is the coefficient $c(f_i, m_j)$ of the polynomial $f_i \in F$ at the monomial $m_j \in U_F$. Thus,
\begin{equation}
M(F)\, \ve{U_F} = \mathbf 0
\end{equation}
is the vector form of the system of polynomial equations $f_1 = \ldots = f_s = 0$. In the following we will refer to such a system as $F = 0$ for short.

A \emph{shift} of a polynomial $f$ is a multiple of $f$ by a monomial $m \in U$. Let $A = (A_1, \ldots, A_s)$ be an ordered $s$-tuple of finite sets of monomials $A_j \subset U$ for all $j$. We define the \emph{set of shifts} of $F$ as
\begin{equation}
\label{eq:AF}
A\cdot F = \{m \cdot f_j \cln \forall m \in A_j, \forall f_j \in F\}.
\end{equation}

Let $a$ be a Laurent monomial ($a = \frac{m_1}{m_2}$, where $m_1, m_2 \in U$ and $a \neq 1$) and $\mathcal B$ be a finite subset of monomials from $U_{A\cdot F}$ such that $a m \in U_{A\cdot F}$ for each $m \in \mathcal B$. The subset $\mathcal B$ is called the set of \emph{basic} monomials. The subsets
\begin{equation}
\label{eq:RE}
\mathcal R = \{a m \cln m \in \mathcal B\} \setminus \mathcal B \quad\text{and}\quad
\mathcal E = U_{A\cdot F} \setminus (\mathcal R \cup \mathcal B)
\end{equation}
are called the sets of \emph{reducible} and \emph{excessive} monomials, respectively~\cite{byrod2009fast}. Clearly, these subsets are pairwise disjoint and $U_{A\cdot F} = \mathcal E \cup \mathcal R \cup \mathcal B$.

A Macaulay matrix $M(A\cdot F)$ with columns arranged in ordered blocks
$
M(A \cdot F) =
\begin{bmatrix}
M_{\mathcal E} & M_{\mathcal R} & M_{\mathcal B}
\end{bmatrix}
$
is called the \emph{elimination template} for $F$ with respect to $a$ if the reduced row echelon form of $M(A\cdot F)$, denoted by $\widetilde M(A\cdot F)$, has the form 
\begin{equation}
\label{eq:Mtilde}
\widetilde M(A\cdot F) =
\hspace{-6pt}
\kbordermatrix{
& \mathcal E & \mathcal R & \mathcal B\\
& * & 0 & * \\
& 0 & I & \widetilde M_{\mathcal B} \\ 
& 0 & 0 & 0},
\end{equation}
where $*$ means a submatrix with arbitrary entries, $0$ is the zero matrix of a suitable size, $I$ is the identity matrix of order $\# \mathcal R$ and $\widetilde M_{\mathcal B}$ is a matrix of size $\#\mathcal R \times \#\mathcal B$.

The elimination template is uniquely determined (up to reordering the rows and columns in the blocks $M_{\mathcal E}$, $M_{\mathcal R}$, $M_{\mathcal B}$) by the following data:
\begin{itemize}
\item the \emph{action} Laurent monomial $a$;
\item the $s$-tuple of sets $A$;
\item the set of basic monomials $\mathcal B$.
\end{itemize}

Knowing the elimination template transforms the root-finding problem for the 0-dimensional polynomial system $F = 0$ into a computationally tractable generalized eigenvalue problem for a pair of square matrices of order $\#\mathcal B$. We will see in Section~\ref{sec:alg} how this works for the SGP forward kinematics problem.

An elimination template can be systematically constructed from a Gr\"obner basis of the polynomial ideal $\langle F\rangle$ with respect to a certain monomial ordering. In this case, $\mathcal B$ represents the standard linear basis of the quotient ring $\mathbb C[X]/\langle F\rangle$ and $\#\mathcal B$ equals the number of roots (with multiplicities) of the 0-dimensional system $F = 0$~\cite{Cox-IVA-2015}. Such an approach has been realized in the automatic template generators of~\cite{kukelova2008automatic,larsson2017efficient,larsson2018beyond,martyushev2022optimizing}. However, Gr\"obner bases are not the only way to construct elimination templates. Recent advances have developed alternative heuristic procedures~\cite{larsson2018beyond,bhayani2020sparse,martyushev2024automatic}. In this case, $\mathcal B$ is not necessarily standard, nor is it necessarily a basis of the quotient ring. For many hard polynomial systems, these heuristic approaches frequently outperform the Gr\"obner-based method in template size reduction.

\section{Problem statement}
\label{sec:prob}

Consider a general Stewart--Gough platform, see Figure~\ref{fig:sgp}. Let $(R, \mathbf t) \in \mathrm{SE}(3)$ be a rigid-body transformation that relates the coordinate frames associated with the base and top platforms. Let $\mathbf x_i \in \mathbb R^3$ and $\mathbf X_i \in \mathbb R^3$ denote the positions of the $i$th attachment point on the top and base platforms, respectively, expressed in their local frames. Let $L_i = l_i^2$, where $l_i$ is the length of the $i$th connecting leg. Then, for each $i = 1, \ldots, 6$, the fixed-distance condition for the $i$th leg yields
\begin{equation}
\label{eq:kinemat}
\|R \mathbf x_i + \mathbf t - \mathbf X_i\|^2 - L_i = 0,
\end{equation}
where $\|\cdot\|$ is the Euclidean (Frobenius) norm.

Let the rotation matrix $R$ be parametrized by the Cayley transform
\begin{equation}
\label{eq:cayley}
R = (I - [\mathbf p]_\times)\, (I + [\mathbf p]_\times)^{-1},
\end{equation}
where $I$ is the $3\times 3$ identity matrix, $\mathbf p \in \mathbb R^3$ is the parameter vector, and $[\mathbf p]_\times$ denotes the skew-symmetric matrix such that $[\mathbf p]_\times \mathbf q = \mathbf p \times \mathbf q$ for any $\mathbf q \in \mathbb R^3$. Substituting~\eqref{eq:cayley} into Eq.~\eqref{eq:kinemat} and clearing denominators yields a system of polynomial equations. The set of the related polynomials is denoted by $F = \{f_1, \ldots, f_6\}$.

The polynomial system $F = 0$ is generically 0-dimensional, admitting exactly 40 complex roots in accordance with the well-known result that the maximum number of solutions (whether complex or real) to the forward kinematics of a general SGP is 40~\cite{raghavan1993stewart,lazard1993representation,mourrain199340,wampler1996forward}. Notably, there exist special instances of $F$ where all 40 solutions are real, distinct, and well separated~\cite{dietmaier1998stewart}.

Through a proper choice of coordinate frames (setting $\mathbf x_1 = \mathbf X_1 = \mathbf 0$), the first polynomial in $F$ simplifies to
\begin{equation}
\label{eq:f1}
f_1 = \|\mathbf t\|^2 - L_1.
\end{equation}
Thus, for each particular values of
\begin{equation}
\begin{split}
&\mathbf x_i,\, \mathbf X_i, \quad &i = 2, \ldots, 6,\\
&L_i, \quad &i = 1, \ldots, 6,
\end{split}
\end{equation}
the problem is to solve for $(\mathbf p, \mathbf t) \in \mathbb R^3 \times \mathbb R^3$ satisfying the six polynomial equations $F = 0$. The first quadratic equation is purely translation-dependent, while the remaining five quartic equations couple both rotation and translation parameters.

\section{Description of the algorithm}
\label{sec:alg}

In this paper we propose to solve the system $F = 0$ by the method of elimination templates. We applied state-of-the-art publicly available automatic template generators from \cite{kukelova2008automatic,larsson2017efficient,martyushev2022optimizing,martyushev2024automatic} to $F$. While the generators from~\cite{kukelova2008automatic,larsson2017efficient,martyushev2022optimizing} failed to construct a template in a reasonable time, the generator from~\cite{martyushev2024automatic} successfully constructed a template of size $293\times 362$. To ensure reproducibility, the complete template construction data is provided in the Appendix. Here we report only the cardinalities of
\begin{itemize}
\item the monomial shift sets: $\#A_1 = 218$, $\#A_2 = 61$, $\#A_3 = 60$, $\#A_4 = 59$, $\#A_5 = 57$, $\#A_6 = 56$;
\item the set of basic monomials: $\#\mathcal B = 69$.
\end{itemize}

The proposed algorithm consists of three main steps: 1) Template construction; 2) PLU decomposition; 3) QZ decomposition. Each step is detailed below.

\subsection{Template construction}

\begin{figure}[ht]
\centering
\includegraphics[width=0.7\textwidth]{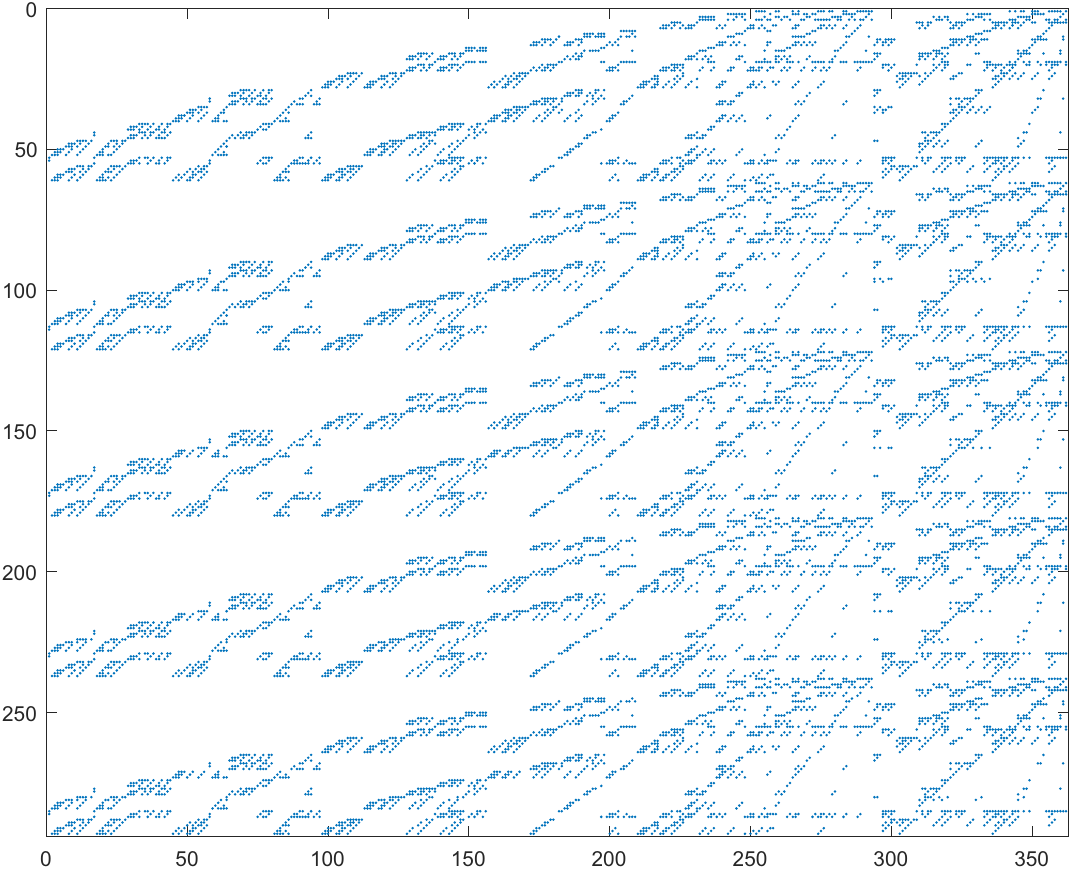}
\caption{Sparse pattern of the elimination template $\widehat M$: blue dots represent non-zero elements}
\label{fig:template}
\end{figure}

The set of shifts of $F$, denoted by $A\cdot F$, is constructed according to Eq.~\eqref{eq:AF}, where $A = (A_1, \ldots, A_6)$ is the 6-tuple of monomial shift sets. This yields an initial Macaulay matrix (elimination template) of size $511\times 580$ with the block structure
\begin{equation}
\label{eq:M}
M(A\cdot F) = \begin{pmatrix}M_{11} & M_{12}\\ M_{21} & M_{22}\end{pmatrix},
\end{equation}
where the submatrix $\begin{pmatrix}M_{11} & M_{12}\end{pmatrix}$ encodes the $218$ shifts of $f_1$. By Schur complement reduction, the template $M(A\cdot F)$ is reduced to the following template of size $293\times 362$:
\begin{equation}
\label{eq:Mhat}
\widehat M = M_{22} - M_{21} M_0,
\end{equation}
where $M_0 = M_{11}^{-1} M_{12}$. The matrix $M_0$ can be safely precomputed offline due to the special sparse form~\eqref{eq:f1} of $f_1$. The resulting template $\widehat M$, visualized in Figure~\ref{fig:template}, has a sparsity of $90\%$.

\subsection{PLU decomposition}

The cardinalities of the sets of basic, reducible, and excessive monomials are $\#\mathcal B = 69$, $\#\mathcal R = 38$, and $\#\mathcal E = 255$ respectively. The monomials in vector $\ve{\mathcal B}$ are ordered so that its last $7$ elements are
\begin{equation}
\label{eq:last7B}
(\mathbf p)_1 (\mathbf p)_3,\quad (\mathbf p)_2 (\mathbf p)_3,\quad (\mathbf p)_3^2,\quad (\mathbf t)_1 (\mathbf p)_3,\quad (\mathbf t)_2 (\mathbf p)_3,\quad (\mathbf t)_3 (\mathbf p)_3,\quad (\mathbf p)_3,
\end{equation}
where $(\mathbf p)_i$ denotes the $i$th entry of vector $\mathbf p$. The template $\widehat M$ has the block structure
$
\widehat M =
\begin{bmatrix}
\widehat M_{\mathcal E} & \widehat M_{\mathcal R} & \widehat M_{\mathcal B}
\end{bmatrix},
$
where $\widehat M_{\mathcal E}$ is a $293\times 255$ sparse matrix. Performing LU decomposition with partial pivoting on $\widehat M_{\mathcal E}$ yields $\widehat M_{\mathcal E} = PLU$, where $P$ is a permutation matrix. Then we have
\begin{equation}
\label{eq:PLM}
(PL)^{-1} \widehat M =
\begin{bmatrix}
* & * & * \\
0 & A_{\mathcal R} & A_{\mathcal B}
\end{bmatrix},
\end{equation}
where $A_{\mathcal R}$ is a full-rank $38 \times 38$ matrix and $A_{\mathcal B}$ is a $38 \times 69$ matrix.

\subsection{QZ decomposition}

It follows from~\eqref{eq:PLM} that $A_{\mathcal R} \ve{\mathcal R} = - A_{\mathcal B} \ve{\mathcal B}$, which translates into the following generalized eigenvalue problem:
\begin{equation}
\label{eq:T0T1}
a T_1 \ve{\mathcal B} = T_0 \ve{\mathcal B},
\end{equation}
where $a = \frac{1}{(\mathbf p)_3}$ is the action Laurent monomial. The $69\times 69$ matrices $T_0$ and $T_1$ are constructed from $A_{\mathcal R}$ and $A_{\mathcal B}$ as follows. Let $b_i$ be the $i$th element of $\ve{\mathcal B}$ and $r_j$ be the $j$th element of $\ve{\mathcal R}$.
\begin{itemize}
\item If $a b_i = r_j$, then
\begin{itemize}
\item $(T_0)_{i,k} = -(A_{\mathcal B})_{j,k}$ for all $k$;
\item $(T_1)_{i,k} = (A_{\mathcal R})_{j,l}$ for all pairs $(k, l)$ such that $a b_k = r_l$.
\end{itemize}
\item If $a b_i = b_j$, then
\begin{itemize}
\item $(T_0)_{i,j} = 1$;
\item $(T_1)_{i,i} = 1$.
\end{itemize}
\end{itemize}
All other entries of matrices $T_0$ and $T_1$ are zero.

The numerically backward stable QZ (generalized Schur) decomposition algorithm~\cite{golub2013matrix,kressner2005numerical} is used to compute the generalized eigenvectors of the matrix pair $(T_0, T_1)$. Let $\mathbf u_i$ denote the $i$th generalized eigenvector, with $(\mathbf u_i)_k$ representing its $k$th element. Then, according to~\eqref{eq:last7B}, the $i$th solution to $F = 0$ is given by
\begin{equation}
\label{eq:pt}
\mathbf p_{e,i} = \frac{1}{(\mathbf u_i)_{69}} \begin{bmatrix}(\mathbf u_i)_{63}\\ (\mathbf u_i)_{64}\\ (\mathbf u_i)_{65}\end{bmatrix}, \quad
\mathbf t_{e,i} = \frac{1}{(\mathbf u_i)_{69}} \begin{bmatrix}(\mathbf u_i)_{66}\\ (\mathbf u_i)_{67}\\ (\mathbf u_i)_{68}\end{bmatrix}.
\end{equation}
The corresponding rotation matrix $R_{e,i}$ is obtained from $\mathbf p_{e,i}$ via the Cayley transform~\eqref{eq:cayley}.

The proposed algorithm outputs $69 - 40 = 29$ false roots that need to be filtered out by back substitution. Let the polynomial system $F = 0$ be written in the form $M(F) \mathbf z = \mathbf 0$, where $M(F)$ and $\mathbf z = \ve{U_F}$ are the normalized Macaulay matrix (with unit row norms) and monomial vector respectively. Let $\mathbf z_i$ be the monomial vector $\mathbf z$ evaluated at the $i$th (possibly false) root. We compute the residuals
\begin{equation}
\label{eq:epsi}
\epsilon_i = \biggl\|M(F)\frac{\mathbf z_i}{\|\mathbf z_i\|}\biggr\|
\end{equation}
and sort them in ascending order. Then the true roots correspond to the first 40 values of $\epsilon_i$.

\section{Experiments}
\label{sec:exper}

The algorithm has been implemented in MATLAB, Julia, and Python. The experiments were conducted on an Intel Core i5-1155G7 processor using the MATLAB implementation. For all the tests, $\mathbf x_1 = \mathbf X_1 = \mathbf 0$ and $\mathbf x_i, \mathbf X_i$ ($i = 2, \ldots, 6$) are uniformly distributed random 3-vectors from $[-2, 2]^{\times 3}$.

While our primary focus is the general 6-6 SGP, the solver also handles two special cases:
\begin{itemize}
\item 6-5: two attachment points on the top platform coincide;
\item 6P-6: semi-planar case (all base attachment points lie in a plane).
\end{itemize}

\begin{remark}
Since the kinematic equations~\eqref{eq:kinemat} are invariant under the transformations $\mathbf x_i \leftrightarrow \mathbf X_i$, $R \rightarrow R^\top$, $\mathbf t \rightarrow -R^\top \mathbf t$, where the superscript $\top$ indicates matrix transpose, the configurations 5-6 and 6-6P are equivalent to 6-5 and 6P-6 respectively.
\end{remark}

\begin{remark}
While the fully planar 6P-6P configuration is a degenerate case for the proposed solver, experiments confirm that the solver is able to handle the ``almost planar'' case, where only one point on the top or base platform is outside the plane.
\end{remark}

\subsection{Numerical accuracy, number of real roots, and speed}

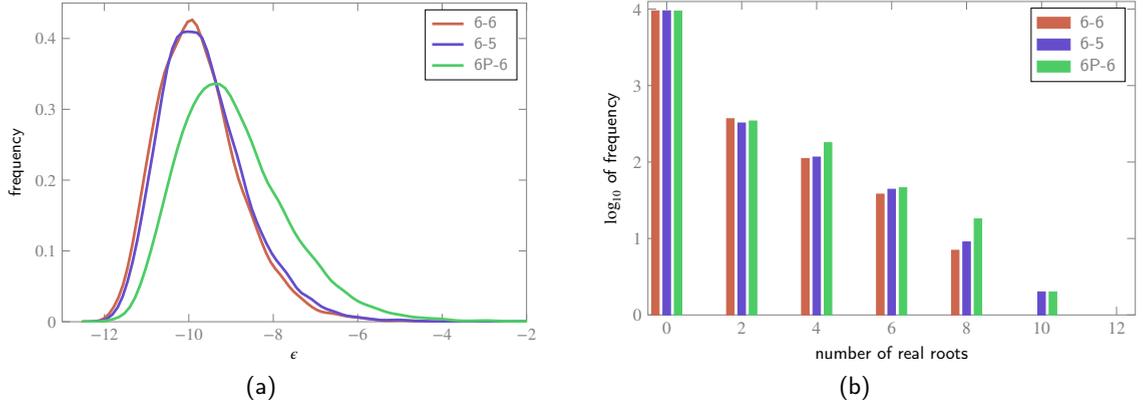
\begin{figure}[ht]
\centering
\begin{tabular}{cc}
\resizebox{0.45\textwidth}{!}{\input{numerror}} & \resizebox{0.45\textwidth}{!}{\input{numrealroots}}\\
(a) & (b)
\end{tabular}
\caption{Distributions over $10^4$ trials: (a) numerical errors, (b) real root counts (logarithmic $y$-axis scale)}
\label{fig:numerr}
\end{figure}

For this test, each squared leg length $L_i$ was sampled independently from a uniform distribution over the interval $[0.5, 3]$. The numerical accuracy is quantified by computing the error metric
\begin{equation}
\label{eq:eps}
\epsilon = \frac{1}{2}\log_{10}\sum_{i = 1}^{40} \epsilon_i^2,
\end{equation}
where the residuals $\epsilon_i$ are defined in Eq.~\eqref{eq:epsi}. Figure~\ref{fig:numerr}(a) shows the error distribution histograms, while Table~\ref{tab:numerr} provides detailed statistics including the median, mean, and maximum values of $\epsilon$ over all trials.

Figure~\ref{fig:numerr}(b) shows the distribution of real root counts using a logarithmic $y$-axis for better visualization. The results reveal that approximately $95\%$ of trials produce no real roots, while the remaining trials produce between $2$ and $10$ real roots. Although the general 6-6 configuration is known to admit a maximum of $40$ real roots~\cite{dietmaier1998stewart}, the maximum number of real roots for both 6-5 and 6P-6 configurations is currently unknown (to the best of the author's knowledge). The observed maximum over more than $10^5$ trials was $12$ for both special cases.

\begin{table}[ht]
\centering
\begin{tabular}{lccccccc}
\hline\\[-6pt]
SGP type & Med. $\epsilon$ & Mean $\epsilon$ & Max. $\epsilon$ & $\epsilon > -5$ & $\epsilon > -6$ & Med. $(\epsilon_{40} - \epsilon_{41})$ & Mean $(\epsilon_{40} - \epsilon_{41})$ \\
\hline\\[-6pt]
6-6 & $-9.86$ & $-9.71$ & $-3.41$ & $0.13\%$ & $0.38\%$ & $-8.69$ & $-8.56$ \\
6-5 & $-9.76$ & $-9.63$ & $-1.34$ & $0.15\%$ & $0.46\%$ & $-8.48$ & $-8.36$ \\
6P-6 & $-9.12$ & $-8.94$ & $-0.71$ & $0.91\%$ & $2.71\%$ & $-7.78$ & $-7.61$ \\
\hline
\end{tabular}
\caption{Complementary to Figure~\ref{fig:numerr}(a). Columns 2--6: error distribution statistics, columns 7--8: true/false roots boundary sharpness statistics}
\label{tab:numerr}
\end{table}

Table~\ref{tab:time} summarizes the average runtime distribution across the three main steps of the algorithm: 1) construction of the elimination template $\widehat M$, see Eq.~\eqref{eq:Mhat}; 2) PLU decomposition of the $293\times 255$ sparse matrix $\widehat M_{\mathcal E}$ followed by matrix division in Eq.~\eqref{eq:PLM}; 3) QZ decomposition of the matrix pair $(T_0, T_1)$. It can be seen that most of the time ($>67\%$) is spent on step~2. Leveraging the specific sparse pattern of the matrix $\widehat M$, as shown in Figure~\ref{fig:template}, could potentially lead to enhanced efficiency in the PLU decomposition step.

\begin{table}[ht]
\centering
\begin{tabular}{lcccc}
\hline\\[-6pt]
Step & Template & PLU & QZ & Total \\
\hline\\[-6pt]
Time (ms) & $0.4$ & $4.2$ & $1.6$ & $6.2$ \\
\hline\\[-6pt]
\end{tabular}
\caption{Average runtime across the main steps of the algorithm}
\label{tab:time}
\end{table}

\subsection{Behaviour under uncertainties}

In this subsection we demonstrate that the solver maintains continuous dependence on input parameters --- small measurement errors produce small deviations in the estimated rigid-body transformation. This stability property ensures predictable behaviour in real-world applications.

For this test, each squared leg length $L_i$ was computed from Eq.~\eqref{eq:kinemat} using standard normally distributed translation ($\mathbf t$) and rotation ($\mathbf p$) vectors. To simulate measurement uncertainty, each squared length was perturbed as $L_i(1 + s_i)^2$, where $s_i$ has a normal distribution with zero mean and standard deviation $\sigma$ ranging from $0$ to $8\times 10^{-4}$.

To evaluate solver accuracy, we compared the estimated rigid-body transformations to the ground truth by computing the rotational and translational errors:
\begin{align}
\label{eq:epsRt}
\epsilon_R &= \min_i\arccos\frac{\tr(R_{\mathrm e,i}^\top R_{\mathrm{gt}}) - 1}{2},\\
\epsilon_{\mathbf t} &= \min_i\arccos\frac{\mathbf t_{\mathrm e,i}^\top \mathbf t_{\mathrm{gt}}}{\|\mathbf t_{\mathrm e,i}\|\, \|\mathbf t_{\mathrm{gt}}\|}.
\end{align}
Here, $\tr(\cdot)$ is the trace function, $(R_{\mathrm{gt}}, \mathbf t_{\mathrm{gt}})$ is the ground truth transformation, $(R_{\mathrm e,i}, \mathbf t_{\mathrm e,i})$ is the $i$th estimated transformation, and the minimum is taken over all real solutions.

\begin{figure}[ht]
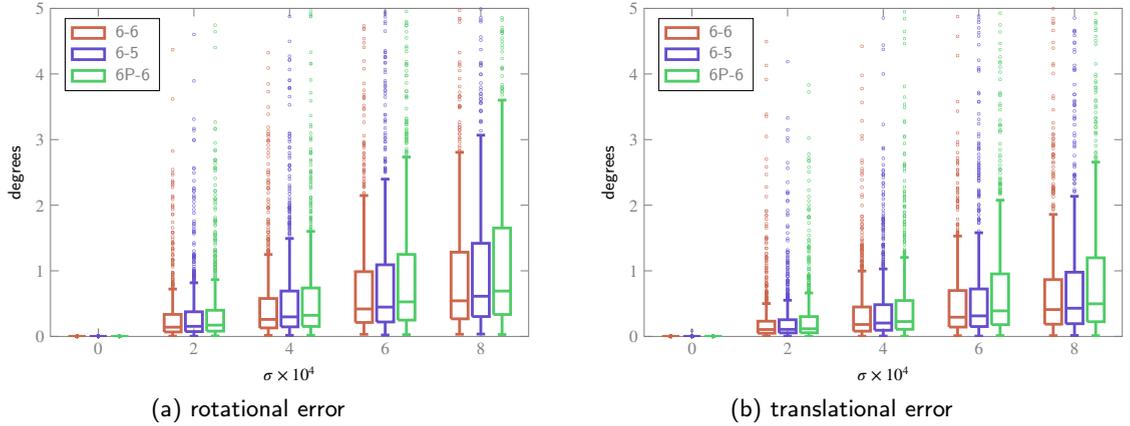

\centering
\begin{tabular}{cc}
\resizebox{0.45\textwidth}{!}{\input{errR}} & \resizebox{0.45\textwidth}{!}{\input{errT}}\\
(a) rotational error & (b) translational error
\end{tabular}
\caption{Rotational and translational errors against the level of leg length measurement uncertainty. Each boxplot summarizes the results from $10^3$ independent trials}
\label{fig:errRT}
\end{figure}

Figure~\ref{fig:errRT} demonstrates the behaviour of the solver as $\sigma$ increases. Each boxplot displays the data distribution with the lower ($25\%$) and upper ($75\%$) quartiles forming the box, the median marked by an internal line, and whiskers extending to data points within $1.5\times$ the interquartile range. Outliers beyond the whiskers are represented by individual dots. The results confirm the predictable solver behaviour for all three SGP types, although the results are slightly worse for the semi-planar case. Under noise-free conditions ($\sigma = 0$), both rotational and translational errors vanish with no outliers observed.

\section{Conclusion}
\label{sec:concl}

This paper proposes a novel practical algorithm for solving the forward kinematics problem for a general Stewart–-Gough platform. The proposed algorithm is purely algebraic, requiring no geometric assumptions about the base or top platforms and no additional sensor data. All 40 solutions to the problem (real and complex) are derived from the generalized eigenvectors of a pair of $69\times 69$ matrices.

A series of experiments on synthetic data validate the algorithm's numerical accuracy and predictable behaviour under measurement uncertainty in leg lengths. With an average runtime of approximately 6ms on standard hardware, the solver is compatible with real-time control for some applications.

Finally, while this paper addresses the general 6-6 SGP configuration along with the 6-5 and 6P-6 special cases, the method of elimination templates provides a general framework for developing efficient solvers for other common types of SGPs, including 6-4, 6-3, 5-5, 5-4, etc.

\appendix

\section{Data for template construction}
\label{app:data}

\setlength\arraycolsep{2.8pt}

This appendix provides the data needed to construct the elimination template found by the automatic generator from~\cite{martyushev2024automatic}. Let $\mathbf p = \begin{bmatrix}u & v & w\end{bmatrix}^\top$ and $\mathbf t = \begin{bmatrix}x & y & z\end{bmatrix}^\top$, see Section~\ref{sec:prob} for description. Then
\begin{itemize}
\item the action Laurent monomial: $a = \frac{1}{w}$;

\item the monomial shift sets:
\[
\begin{array}{llllllllllllll}
A_1 & = & \{1, & z, & y, & x, & w, & v, & u, & z^2, & yz, & xz, & zw, & vz,\\
& & uz, & y^2, & xy, & wy, & vy, & uy, & x^2, & wx, & vx, & ux, & w^2, & vw,\\
& & uw, & v^2, & uv, & u^2, & zwy, & yvz, & yuz, & zwx, & xvz, & zxu, & zw^2, & zv^2,\\
& & zvu, & zu^2, & y^2w, & y^2v, & y^2u, & yxw, & yxv, & yxu, & yw^2, & ywu, & yv^2, & yvu,\\
& & yu^2, & wx^2, & vx^2, & ux^2, & xw^2, & xwv, & xwu, & xv^2, & xvu, & xu^2, & w^3, & w^2v,\\
& & w^2u, & wv^2, & wvu, & wu^2, & v^3, & v^2u, & vu^2, & u^3, & w^2z^2, & v^2z^2, & u^2z^2, & yzw^2,\\
& & yzv^2, & yzu^2, & xzw^2, & xzv^2, & xzu^2, & w^3z, & vzw^2, & uzw^2, & v^2zw, & u^2zw, & v^3z, & uzv^2,\\
& & vzu^2, & u^3z, & w^2y^2, & v^2y^2, & u^2y^2, & xyw^2, & xyv^2, & xyu^2, & w^3y, & vyw^2, & uyw^2, & v^2yw,\\
& & wyu^2, & v^3y, & uyv^2, & u^2yv, & u^3y, & w^2x^2, & v^2x^2, & u^2x^2, & w^3x, & vxw^2, & uxw^2, & wxv^2,\\
& & u^2xw, & v^3x, & uxv^2, & u^2xv, & u^3x, & w^4, & w^3v, & w^3u, & w^2v^2, & w^2uv, & w^2u^2, & v^3w,\\
& & v^2uw, & vu^2w, & u^3w, & v^4, & v^3u, & v^2u^2, & u^3v, & u^4, & zw^3y, & zvyw^2, & zuyw^2, & zv^2yw,\\
& & zwyu^2, & zv^3y, & zuyv^2, & zu^2yv, & zu^3y, & zw^3x, & zvxw^2, & zuxw^2, & zwxv^2, & zu^2xw, & zv^3x, & zuxv^2,\\
& & zu^2xv, & zu^3x, & w^4z, & vw^3z, & uw^3z, & v^2w^2z, & uvw^2z, & u^2w^2z, & v^3wz, & uv^2wz, & u^2vwz, & u^3wz,\\
& & v^4z, & uv^3z, & u^2v^2z, & u^3vz, & u^4z, & w^3y^2, & vw^2y^2, & uw^2y^2, & wv^2y^2, & wu^2y^2, & v^3y^2, & uv^2y^2,\\
& & vu^2y^2, & u^3y^2, & w^3xy, & vw^2xy, & uw^2xy, & v^2wxy, & u^2wxy, & v^3xy, & uv^2xy, & u^2vxy, & u^3xy, & w^4y,\\
& & vw^3y, & uw^3y, & v^2w^2y, & uvw^2y, & u^2w^2y, & v^3wy, & uv^2wy, & u^2vwy, & u^3wy, & v^4y, & uv^3y, & u^2v^2y,\\
& & u^3vy, & u^4y, & w^3x^2, & vw^2x^2, & uw^2x^2, & wv^2x^2, & wu^2x^2, & v^3x^2, & uv^2x^2, & vu^2x^2, & u^3x^2, & w^4x,\\
& & vw^3x, & uw^3x, & v^2w^2x, & uvw^2x, & u^2w^2x, & v^3wx, & uv^2wx, & u^2vwx, & u^3wx, & v^4x, & uv^3x, & u^2v^2x,\\
& & u^3vx, & u^4x\}, & & & & & & & & & &
\end{array}
\]
\[
\begin{array}{lllllllllllllllll}
A_2 & = & \{1, & z, & y, & x, & w, & v, & u, & z^2, & yz, & xz, & zw, & vz, & uz, & y^2, & xy,\\
& & wy, & vy, & uy, & x^2, & wx, & vx, & ux, & w^2, & vw, & uw, & v^2, & uv, & u^2, & zwy, & yvz,\\
& & yuz, & zwx, & xvz, & zxu, & zw^2, & zwv, & zwu, & zv^2, & zvu, & zu^2, & y^2w, & y^2v, & y^2u, & yxw, & yxv,\\
& & yxu, & yw^2, & ywv, & ywu, & yv^2, & yvu, & yu^2, & wx^2, & vx^2, & ux^2, & xw^2, & xwv, & xwu, & xv^2, & xvu,\\
& & xu^2\}, & & & & & & & & & & & & & & \\[5pt]
A_3 & = & \{1, & z, & y, & x, & w, & v, & u, & z^2, & yz, & xz, & zw, & vz, & uz, & y^2, & xy,\\
& & wy, & vy, & uy, & x^2, & wx, & vx, & ux, & w^2, & vw, & uw, & v^2, & uv, & u^2, & zwy, & yvz,\\
& & yuz, & zwx, & xvz, & zxu, & zw^2, & zwu, & zv^2, & zvu, & zu^2, & y^2w, & y^2v, & y^2u, & yxw, & yxv, & yxu,\\
& & yw^2, & ywv, & ywu, & yv^2, & yvu, & yu^2, & wx^2, & vx^2, & ux^2, & xw^2, & xwv, & xwu, & xv^2, & xvu, & xu^2\}, \\[5pt]
A_4 & = & \{1, & z, & y, & x, & w, & v, & u, & z^2, & yz, & xz, & zw, & vz, & uz, & y^2, & xy,\\
& & wy, & vy, & uy, & x^2, & wx, & vx, & ux, & w^2, & vw, & uw, & v^2, & uv, & u^2, & zwy, & yvz,\\
& & yuz, & zwx, & xvz, & zxu, & zw^2, & zv^2, & zvu, & zu^2, & y^2w, & y^2v, & y^2u, & yxw, & yxv, & yxu, & yw^2,\\
& & ywv, & ywu, & yv^2, & yvu, & yu^2, & wx^2, & vx^2, & ux^2, & xw^2, & xwv, & xwu, & xv^2, & xvu, & xu^2\}, & \\[5pt]
A_5 & = & \{1, & z, & y, & x, & w, & v, & u, & yz, & xz, & zw, & vz, & uz, & y^2, & xy, & wy,\\
& & vy, & uy, & x^2, & wx, & vx, & ux, & w^2, & vw, & uw, & v^2, & uv, & u^2, & zwy, & yvz, & yuz,\\
& & zwx, & xvz, & zxu, & zw^2, & zv^2, & zvu, & zu^2, & y^2w, & y^2v, & y^2u, & yxw, & yxv, & yxu, & yw^2, & ywu,\\
& & yv^2, & yvu, & yu^2, & wx^2, & vx^2, & ux^2, & xw^2, & xwv, & xwu, & xv^2, & xvu, & xu^2\}, & & & \\[5pt]
A_6 & = & \{1, & z, & y, & x, & w, & v, & u, & yz, & xz, & zw, & vz, & uz, &
y^2, & xy, & wy,\\
& & vy, & uy, & x^2, & wx, & vx, & ux, & w^2, & vw, & uw, & v^2, & uv, & u^2, & zwy, & yvz, & yuz,\\
& & zwx, & xvz, & zxu, & zv^2, & zvu, & zu^2, & y^2w, & y^2v, & y^2u, & yxw, & yxv, & yxu, & yw^2, & ywu, & yv^2,\\
& & yvu, & yu^2, & wx^2, & vx^2, & ux^2, & xw^2, & xwv, & xwu, & xv^2, & xvu, & xu^2\}; & & & &
\end{array}
\]

\item the set of basic monomials:
\[
\begin{array}{llllllllllllll}
\mathcal B & = & \{zuyw^2, & zvyw^2, & zw^3y, & z^2uwv, & wv^2z^2, & uw^2z^2, & vw^2z^2, & w^3z^2, & v^3w, & w^2u^2, & w^2uv, & w^2v^2,\\
& & w^3u, & w^3v, & w^4, & w^3x, & uywv, & v^2yw, & uyw^2, & vyw^2, & w^3y, & yxwu, & xywv, & xyw^2,\\
& & uwy^2, & y^2wv, & w^2y^2, & u^2zw, & uzwv, & v^2zw, & uzw^2, & vzw^2, & w^3z, & xzwu, & xzwv, & xzw^2,\\
& & yzwu, & yzwv, & yzw^2, & z^2wu, & z^2wv, & w^2z^2, & wu^2, & wvu, & wv^2, & w^2u, & w^2v, & w^3,\\
& & xwu, & xwv, & xw^2, & ywu, & ywv, & yw^2, & yxw, & wy^2, & zwu, & zwv, & zw^2, & zwx,\\
& & zwy, & wz^2, & wu, & vw, & w^2, & xw, & yw, & zw, & w\}. & & &
\end{array}
\]
\end{itemize}


\printcredits

\section*{Declaration of competing interest}

The author declares that he has no known competing financial interests or personal relationships that could appear to influence the work in this paper.

\section*{Data availability}

Reference implementations of the proposed algorithm in MATLAB, Julia, and Python are available under an open-source license at
\href{https://github.com/martyushev/fkSGP}{https://github.com/martyushev/fkSGP}.

\bibliographystyle{cas-model2-names}

\bibliography{biblio}

\end{document}

%% file: numerror.tex
%
%
\definecolor{mycolor2}{rgb}{0.30000,0.80000,0.40000}
\definecolor{mycolor1}{rgb}{0.40000,0.30000,0.80000}
\definecolor{mycolor3}{rgb}{0.80000,0.40000,0.30000}

\newcommand{\lw}{1.5pt}

\begin{tikzpicture}

\begin{axis}[%
width=3.5in,
height=2.4in,
at={(1.85in,1.215in)},
scale only axis,
xmin=-13,
xmax=-2,
xtick={-12,-10,-8,-6,-4,-2},
xticklabels={$-12$,$-10$,$-8$,$-6$,$-4$,$-2$},
xlabel style={font=\color{black}},
xlabel={$\epsilon$},
ymin=0,
ymax=0.45,
ylabel style={font=\color{black}},
ylabel={frequency},
legend style={legend cell align=left, align=left, draw=black}
]
\addplot [color=mycolor3, line width=\lw]
  table[row sep=crcr]{%
-12.540170156844	3.41486868756335e-06\\
-12.4309570357342	2.62584789652172e-05\\
-12.3217439146243	0.000144700553523047\\
-12.2125307935145	0.000597000788188461\\
-12.1033176724047	0.00195385139600086\\
-11.9941045512949	0.00521629299378772\\
-11.8848914301851	0.011454649789581\\
-11.7756783090753	0.0210991287912696\\
-11.6664651879655	0.03429692338963\\
-11.5572520668556	0.052321767925865\\
-11.4480389457458	0.0769080412816472\\
-11.338825824636	0.108071994108623\\
-11.2296127035262	0.144070061125814\\
-11.1203995824164	0.182437543063416\\
-11.0111864613066	0.221058989770231\\
-10.9019733401968	0.258840441289996\\
-10.7927602190869	0.29477496964366\\
-10.6835470979771	0.326679325227271\\
-10.5743339768673	0.351863424852061\\
-10.4651208557575	0.369850559581467\\
-10.3559077346477	0.383701897280539\\
-10.2466946135379	0.397642858745562\\
-10.1374814924281	0.412312498851112\\
-10.0282683713182	0.423478179010507\\
-9.91905525020843	0.426270318974636\\
-9.80984212909861	0.418250182659441\\
-9.7006290079888	0.400969847005634\\
-9.59141588687898	0.380516698346137\\
-9.48220276576917	0.361140968116267\\
-9.37298964465935	0.340027337266391\\
-9.26377652354954	0.313450667608183\\
-9.15456340243973	0.28318271170045\\
-9.04535028132991	0.253310062881686\\
-8.9361371602201	0.225931545297964\\
-8.82692403911028	0.20238932776989\\
-8.71771091800047	0.183392919945247\\
-8.60849779689065	0.166619233646654\\
-8.49928467578084	0.149161548219173\\
-8.39007155467102	0.131053704690286\\
-8.28085843356121	0.1135681535921\\
-8.17164531245139	0.0975594164503668\\
-8.06243219134158	0.0844577761748183\\
-7.95321907023177	0.075032652318331\\
-7.84400594912195	0.0672737043858254\\
-7.73479282801214	0.0587882563436619\\
-7.62557970690232	0.0499869476506912\\
-7.51636658579251	0.0426522305439392\\
-7.40715346468269	0.0368301629508737\\
-7.29794034357288	0.0312323299794935\\
-7.18872722246306	0.0256558939793365\\
-7.07951410135325	0.0207426943590275\\
-6.97030098024343	0.0168234981671292\\
-6.86108785913362	0.0141628498892595\\
-6.75187473802381	0.0128229461265848\\
-6.64266161691399	0.0121047539693164\\
-6.53344849580418	0.0111100214931006\\
-6.42423537469436	0.00970712631920276\\
-6.31502225358455	0.00836192688670036\\
-6.20580913247473	0.00731730732155399\\
-6.09659601136492	0.00654294514503853\\
-5.9873828902551	0.00598974245278199\\
-5.87816976914529	0.0054212470469191\\
-5.76895664803547	0.00458389226690464\\
-5.65974352692566	0.00357199934060868\\
-5.55053040581585	0.0026512801874013\\
-5.44131728470603	0.00194704452487033\\
-5.33210416359622	0.00152167872395289\\
-5.2228910424864	0.00145498699164775\\
-5.11367792137659	0.00166459814061028\\
-5.00446480026677	0.00185409150172352\\
-4.89525167915696	0.00177986261822095\\
-4.78603855804714	0.0014699357716982\\
-4.67682543693733	0.00104638911630086\\
-4.56761231582751	0.000607591553511852\\
-4.4583991947177	0.000306356350367774\\
-4.34918607360789	0.000234706798300351\\
-4.23997295249807	0.000307465747775263\\
-4.13075983138826	0.000378390826931061\\
-4.02154671027844	0.000375553975511937\\
-3.91233358916863	0.00029154140239302\\
-3.80312046805881	0.000163433179942999\\
-3.693907346949	6.13680209978089e-05\\
-3.58469422583918	1.47974703690467e-05\\
-3.47548110472937	2.23997240725386e-06\\
-3.36626798361955	2.13643663223538e-07\\
-3.25705486250974	7.17569776275993e-14\\
-3.14784174139993	5.41468819228043e-12\\
-3.03862862029011	2.54571770181384e-10\\
-2.9294154991803	7.45717796681935e-09\\
-2.82020237807048	1.36102545679176e-07\\
-2.71098925696067	1.54769530243142e-06\\
-2.60177613585085	1.09655938415728e-05\\
-2.49256301474104	4.84067942881406e-05\\
-2.38334989363122	0.000133139819831828\\
-2.27413677252141	0.000228158725676591\\
-2.16492365141159	0.000243609194373302\\
-2.05571053030178	0.000162060712499688\\
-1.94649740919196	6.71721559744344e-05\\
-1.83728428808215	1.73471569145889e-05\\
-1.72807116697234	2.79122471235075e-06\\
};
\addlegendentry{~6-6}

\addplot [color=mycolor1, line width=\lw]
  table[row sep=crcr]{%
-12.5120321600264	4.0117313973654e-06\\
-12.3942641005267	3.32103670737243e-05\\
-12.276496041027	0.000179115363290907\\
-12.1587279815274	0.000674925287874848\\
-12.0409599220277	0.00196962922442536\\
-11.923191862528	0.00485091045934415\\
-11.8054238030284	0.0104727977297455\\
-11.6876557435287	0.0201475946752618\\
-11.5698876840291	0.0352607564722019\\
-11.4521196245294	0.0566263467833112\\
-11.3343515650297	0.0837106278244676\\
-11.2165835055301	0.116155276630264\\
-11.0988154460304	0.153885980724715\\
-10.9810473865307	0.19425023595121\\
-10.8632793270311	0.234744768150357\\
-10.7455112675314	0.277035322098563\\
-10.6277432080317	0.320449783922243\\
-10.5099751485321	0.358318070658082\\
-10.3922070890324	0.38549711176814\\
-10.2744390295328	0.401651022366325\\
-10.1566709700331	0.408561996067572\\
-10.0389029105334	0.409639134014853\\
-9.92113485103376	0.409234311285291\\
-9.8033667915341	0.408158774966829\\
-9.68559873203443	0.401033915025264\\
-9.56783067253477	0.383280455316015\\
-9.4500626130351	0.358447436599899\\
-9.33229455353544	0.332754397398438\\
-9.21452649403578	0.307428402926179\\
-9.09675843453611	0.282155548494787\\
-8.97899037503645	0.258762223893265\\
-8.86122231553679	0.235839431996623\\
-8.74345425603712	0.209971825639025\\
-8.62568619653746	0.182949005906093\\
-8.50791813703779	0.159418638645455\\
-8.39015007753813	0.140759337706751\\
-8.27238201803847	0.124749756440851\\
-8.1546139585388	0.109782636064048\\
-8.03684589903914	0.0967083408266964\\
-7.91907783953947	0.0863723933482763\\
-7.80130978003981	0.0777098308490486\\
-7.68354172054015	0.0687667249180051\\
-7.56577366104048	0.0583682878663775\\
-7.44800560154082	0.0479207831565102\\
-7.33023754204115	0.0405996627057938\\
-7.21246948254149	0.0363718778413434\\
-7.09470142304183	0.0319838525431769\\
-6.97693336354216	0.0264544010101796\\
-6.8591653040425	0.0214724002059127\\
-6.74139724454283	0.0180621766368931\\
-6.62362918504317	0.0156955285238957\\
-6.50586112554351	0.0133058994863015\\
-6.38809306604384	0.0106348129418559\\
-6.27032500654418	0.00852268928813536\\
-6.15255694704451	0.00745459594935144\\
-6.03478888754485	0.00667205008286992\\
-5.91702082804519	0.00554315862780874\\
-5.79925276854552	0.00448152255045872\\
-5.68148470904586	0.00375883111607201\\
-5.56371664954619	0.00300196576824038\\
-5.44594859004653	0.00231272589108972\\
-5.32818053054687	0.00217917279097757\\
-5.2104124710472	0.00243565173563116\\
-5.09264441154754	0.00258578482572348\\
-4.97487635204787	0.00246664917246086\\
-4.85710829254821	0.00206021720544513\\
-4.73934023304855	0.00156664386244912\\
-4.62157217354888	0.00127386469653356\\
-4.50380411404922	0.00108394216363252\\
-4.38603605454955	0.000803876360043511\\
-4.26826799504989	0.000461764978248616\\
-4.15049993555023	0.000188519063115755\\
-4.03273187605056	5.0781343209769e-05\\
-3.9149638165509	8.63252826975415e-06\\
-3.79719575705123	2.8811482462467e-06\\
-3.67942769755157	1.46105185715871e-05\\
-3.56165963805191	6.30449047176237e-05\\
-3.44389157855224	0.000160540253554072\\
-3.32612351905258	0.000241250091217547\\
-3.20835545955291	0.000213944249451578\\
-3.09058740005325	0.000111965271900371\\
-2.97281934055359	3.4579259661916e-05\\
-2.85505128105392	6.30247605791534e-06\\
-2.73728322155426	6.84975827187771e-07\\
-2.6195151620546	1.5153731464608e-07\\
-2.50174710255493	1.89985318064904e-06\\
-2.38397904305527	1.40562699789939e-05\\
-2.2662109835556	6.13719406136535e-05\\
-2.14844292405594	0.000158132670068796\\
-2.03067486455628	0.000240474033786075\\
-1.91290680505661	0.000216231659782681\\
-1.79513874555695	0.000119000494833963\\
-1.67737068605728	6.37730821521696e-05\\
-1.55960262655762	0.000104504228154216\\
-1.44183456705796	0.000202301047195689\\
-1.32406650755829	0.000244902179447068\\
-1.20629844805863	0.000175581110983662\\
-1.08853038855896	7.4286949528855e-05\\
-0.970762329059301	1.85479897565749e-05\\
-0.852994269559636	2.73294416774161e-06\\
};
\addlegendentry{~6-5}

\addplot [color=mycolor2, line width=\lw]
  table[row sep=crcr]{%
-12.5269774506619	2.82457703408766e-06\\
-12.4015060958867	1.84386722561477e-05\\
-12.2760347411115	8.75602995031303e-05\\
-12.1505633863363	0.00031490986462154\\
-12.0250920315612	0.000887829446251307\\
-11.899620676786	0.00201950650813765\\
-11.7741493220108	0.0038898611843356\\
-11.6486779672356	0.00685835028908051\\
-11.5232066124604	0.0119109214200675\\
-11.3977352576852	0.0206038011090318\\
-11.27226390291	0.0341436942376099\\
-11.1467925481349	0.0526411845746089\\
-11.0213211933597	0.0753088353402239\\
-10.8958498385845	0.101084620177311\\
-10.7703784838093	0.129013951343345\\
-10.6449071290341	0.158303214905605\\
-10.5194357742589	0.188144886939487\\
-10.3939644194838	0.217391838863302\\
-10.2684930647086	0.244543251497146\\
-10.1430217099334	0.268449160621246\\
-10.0175503551582	0.288778885849249\\
-9.89207900038301	0.305656479425824\\
-9.76660764560782	0.319070183295403\\
-9.64113629083264	0.328606251646447\\
-9.51566493605745	0.334076414001889\\
-9.39019358128227	0.336194852501768\\
-9.26472222650708	0.335256059457523\\
-9.13925087173189	0.329847894220865\\
-9.01377951695671	0.318921501263018\\
-8.88830816218152	0.304139409572424\\
-8.76283680740634	0.28823875906256\\
-8.63736545263115	0.271986189190733\\
-8.51189409785597	0.254072404574044\\
-8.38642274308078	0.233841735311209\\
-8.2609513883056	0.213554150961182\\
-8.13548003353041	0.196721034955268\\
-8.01000867875523	0.183808671175685\\
-7.88453732398004	0.171547061153146\\
-7.75906596920485	0.157263950900259\\
-7.63359461442967	0.141801007776896\\
-7.50812325965448	0.12750504752035\\
-7.3826519048793	0.11543725233592\\
-7.25718055010411	0.105244138489356\\
-7.13170919532893	0.0962408532029512\\
-7.00623784055374	0.0874761163457804\\
-6.88076648577856	0.0777811871231603\\
-6.75529513100337	0.0673407896986027\\
-6.62982377622818	0.0581442031724801\\
-6.504352421453	0.0514058494642788\\
-6.37888106667781	0.0459248935114513\\
-6.25340971190263	0.0401007189763605\\
-6.12793835712744	0.0341382005517966\\
-6.00246700235226	0.029224368680491\\
-5.87699564757707	0.0256706155303247\\
-5.75152429280188	0.0228653850897954\\
-5.6260529380267	0.0202281327439311\\
-5.50058158325151	0.0175031444036383\\
-5.37511022847633	0.0147786113465068\\
-5.24963887370114	0.0125311348995356\\
-5.12416751892596	0.0110255746879143\\
-4.99869616415077	0.00981634284405878\\
-4.87322480937559	0.00845088813335524\\
-4.7477534546004	0.00713830789124785\\
-4.62228209982521	0.00620893188036655\\
-4.49681074505003	0.00557129501599162\\
-4.37133939027484	0.00504882457647835\\
-4.24586803549966	0.00464751676122384\\
-4.12039668072447	0.00429853483493405\\
-3.99492532594929	0.00378461112655699\\
-3.8694539711741	0.003020201963708\\
-3.74398261639892	0.0021777849652029\\
-3.61851126162373	0.00153552876373051\\
-3.49303990684855	0.0012452271125007\\
-3.36756855207336	0.00123422116781628\\
-3.24209719729817	0.00135789664333108\\
-3.11662584252299	0.00154168585760879\\
-2.9911544877478	0.00167551606953491\\
-2.86568313297262	0.00158811936307517\\
-2.74021177819743	0.00126837817674204\\
-2.61474042342225	0.000920195737687914\\
-2.48926906864706	0.000733176255539431\\
-2.36379771387188	0.000715537391386286\\
-2.23832635909669	0.000746556286252755\\
-2.1128550043215	0.000713693895386576\\
-1.98738364954632	0.00059151930776412\\
-1.86191229477113	0.000439179513034391\\
-1.73644093999595	0.000341465550436038\\
-1.61096958522076	0.000328624375714846\\
-1.48549823044558	0.00034953523848179\\
-1.36002687567039	0.000333831679578904\\
-1.23455552089521	0.000269815950540259\\
-1.10908416612002	0.000219608020942735\\
-0.983612811344834	0.000253680298559707\\
-0.858141456569649	0.00034972243748458\\
-0.732670101794463	0.000392268746037683\\
-0.607198747019279	0.000312688803489823\\
-0.481727392244091	0.000172292557716576\\
-0.356256037468906	6.53315556653212e-05\\
-0.23078468269372	1.70169369468661e-05\\
-0.105313327918535	3.04332293624549e-06\\
};
\addlegendentry{~6P-6}

\end{axis}
\end{tikzpicture}%

%% file: numrealroots.tex
%
%
\definecolor{mycolor3}{rgb}{0.30000,0.80000,0.40000}
\definecolor{mycolor2}{rgb}{0.40000,0.30000,0.80000}
\definecolor{mycolor1}{rgb}{0.80000,0.40000,0.30000}

\newcommand{\bw}{0.2}

\begin{tikzpicture}

\begin{axis}[%
width=3.5in,
height=2.25in,
at={(1.85in,1.215in)},
scale only axis,
bar shift auto,
xmin=0.5,
xmax=13.5,
xtick={1,3,5,7,9,11,13},
xticklabels={$0$,$2$,$4$,$6$,$8$,$10$,$12$},
xlabel style={font=\color{black}},
xlabel={number of real roots},
ymin=0,
ymax=4.1,
ylabel style={font=\color{black}},
ylabel={$\log_{10}$ of frequency},
legend style={legend cell align=left, align=left, draw=black}
]
\addplot[ybar, bar width=\bw, fill=mycolor1, draw=mycolor1, area legend] table[row sep=crcr] {%
1	3.97662505205073\\
2	-inf\\
3	2.56584781867352\\
4	-inf\\
5	2.04532297878666\\
6	-inf\\
7	1.57978359661681\\
8	-inf\\
9	0.845098040014257\\
10	-inf\\
11	-inf\\
12	-inf\\
13	-inf\\
};
\addlegendentry{~6-6}

\addplot[ybar, bar width=\bw, fill=mycolor2, draw=mycolor2, area legend] table[row sep=crcr] {%
1	3.97795212120146\\
2	-inf\\
3	2.51054501020661\\
4	-inf\\
5	2.06445798922692\\
6	-inf\\
7	1.64345267648619\\
8	-inf\\
9	0.954242509439325\\
10	-inf\\
11	0.301029995663981\\
12	-inf\\
13	-inf\\
};
\addlegendentry{~6-5}

\addplot[ybar, bar width=\bw, fill=mycolor3, draw=mycolor3, area legend] table[row sep=crcr] {%
1	3.97363577341741\\
2	-inf\\
3	2.53655844257153\\
4	-inf\\
5	2.25285303097989\\
6	-inf\\
7	1.66275783168157\\
8	-inf\\
9	1.25527250510331\\
10	-inf\\
11	0.301029995663981\\
12	-inf\\
13	-inf\\
};
\addlegendentry{~6P-6}

\end{axis}
\end{tikzpicture}%